\documentclass{article} % For LaTeX2e
\usepackage{nips15submit_e,times}
\usepackage{hyperref}
\usepackage{url}
\usepackage{float}
\usepackage{graphicx}
\usepackage{wrapfig}
\usepackage{multicol}
\usepackage{amsmath}
\usepackage{xcolor}
\usepackage{amssymb}
\usepackage{caption}
\usepackage{subcaption}
\usepackage{multirow}
\usepackage{tabularx}
\usepackage{natbib}
\usepackage{amssymb}
\usepackage{paralist}
\usepackage{adjustbox}

\title{Textually Enriched Neural Module Networks for Visual Question Answering}

\author{
Khyathi Chandu, Mary Arpita Pyreddy, Matthieu Felix, Narendra Nath Joshi\\\\
Language Technologies Institute, School of Computer Science\\
Carnegie Mellon University, Pittsburgh, PA\\\\
\texttt{\{kchandu, mpyreddy, matthief, nnj\}@andrew.cmu.edu} \\
}

\DeclareMathOperator{\argmax}{arg\,max}

\nipsfinalcopy % Uncomment for camera-ready version

\begin{document}

\maketitle

\begin{abstract}
Problems at the intersection of language and vision, like visual question answering, have recently been gaining a lot of attention in the field of multi-modal machine learning as computer vision research moves beyond traditional recognition tasks. There has been recent success in visual question answering using deep neural network models which use the linguistic structure of the questions to dynamically instantiate network layouts. In the process of converting the question to a network layout, the question is simplified, which results in loss of information in the model. In this paper, we enrich the image information with textual data using image captions and external knowledge bases to generate more coherent answers. We achieve 57.1\% overall accuracy on the test-dev open-ended questions from the visual question answering (VQA 1.0) real image dataset.
\end{abstract}

\section{Introduction}
% Define the research problem
% Challenges present in VQA

% Research in multi-disciplinary problems such as computer vision combined with natural language processing and knowledge representation allows us to combine multiple modalities in different and interesting ways in an attempt to come a step closer to solving artificial intelligence.

% This research problem is very exciting as it moves beyond the standard specific recognition tasks and employs a variety of AI tasks like object detection, activity recognition, knowledge representation and common-sense reasoning.

% From MT report comments:
% 1st paragraph: motivation, then definition
% 2nd          : prior work and challenges
% 3rd          : contributions/model
% !NO UNDEFINED TERMS

Visual question answering (VQA), introduced by \citet{antol2015vqa}, has generated a lot of interest in recent computer vision research. Given an image and a question in natural language about the image, the task of VQA is to provide an accurate natural language answer (for instance, a simple question could be ``How many people are in the picture?''). Achieving consistently good performance in VQA could have a tremendous impact on both the research community and society in general: it can be used, for instance, for the assistance of the visually impaired for daily navigation, as well as lead us to deeper understanding and better representations for other computer vision tasks like image search. We believe it can have a high impact on multi-modal conversational agent research as well. 

% Not super relevant to what we do: Attention models for VQA by \citet{lu2016hierarchical} and \citet{andreas2016neural}, amongst others, are interesting as they model ``where to look'' in an image when trying to answer the given question.

\citet{andreas2016neural} and \citet{andreas2016learning} propose neural module networks (NMN) for VQA where a network layout is generated by putting together neural modules based on a natural language dependency parse (that is, a tree structure representing relationship between the words) of the question. One limitation in this model is that some of the information present in the question is destroyed when the question is converted to a network layout: NMNs obtain better results with short parse trees, which represent only the main elements of the original question.

In this paper, we propose two approaches to textually enrich NMNs for VQA by leveraging the image captions. The first approach is to incorporate image caption (see \citet{karpathy2015deep}, for instance) information into the model, making the model resilient to information deletion stemming from incorrect or simplified parses. The second approach is a modification of the first approach where we attend to the caption to pick only the useful parts instead of using the caption as a whole. This is to ensure that irrelevant captions do not introduce noise into the system. We also propose a third approach where we leverage information from external knowledge sources to provide better answers to questions that might benefit from additional knowledge. Additionally, we implement the Measure module as proposed by \citet{andreas2016neural}, predominantly for answering yes/no questions.

We first introduce the related work in the next section, propose our approaches and describe our experimental setup, and report our results and discuss our analysis. Finally, we conclude by discussing some possible future directions.

\section{Related Work}
% Describe various techniques in the literature.
% The shortcomings in NMN.
% Last para, how we plan to achieve it.

% The goal of this section is twofold: (1) present an overview of the work happening in your research area, and (2) highlight how your approach differs from prior work.Instead of simply listing all the prior work, try to group them by category or research problem.
%This will allow you to have a more structured related work section and will make it easier for the reader to place your work in the right context. This section should include about 12-15 citations of prior work. Usually, in the last paragraph of this section, you will have a paragraph stating how your work differs from the prior work (e.g., “To our knowledge, this paper is the first to…”).

\subsection{Visual Question Answering}

We review here those that we have deemed most representative of the current approaches in VQA.
\paragraph{Late Fusion} \citet{antol2015vqa}, \citet{malinowski2015ask}, and \citet{gao2015you} use a model where a long short-term memory network (LSTM, \citet{hochreiter1997long}) and a convolutional neural network (CNN, \citet{lecun1989backpropagation}), both pre-trained, run independently on the questions and the images, respectively. The image embedding is transformed to a smaller one (for instance, 1024 dimensions) by a fully-connected layer with $\tanh$ nonlinearity. The output of both networks are fused via element-wise multiplication, and a final fully connected softmax layer is added. \citet{ren2015exploring} perform asymmetric fusion by feeding the output of the CNN into the first LSTM, as though it were the first word of the sentence. This performs somewhat better than naive late fusion.

\paragraph{Attention Models} This family of models assign weights to different parts of the image, in order to filter out irrelevant information that could generate noise in the answer. The model developed by \citet{shih2016where} does this by computing the dot-products of text features extracted from the question and region-by-region features extracted by a CNN. The model then weighs information from each region by the corresponding dot-product value to produce a final answer (information for each region is generated much like in the late-fusion approach).

\citet{lu2016hierarchical} and \citet{nam2016dual} argue that it is equally important to model ``which words to listen to'' (question attention). They present a co-attention model for VQA that jointly performs image and question attention. In addition, their model attends to the question (and consequently the image via the co-attention mechanism) in a hierarchical fashion via a 1-dimensional CNN. \citet{nam2016dual} show that Dual Attention Networks (DANs) jointly leverage visual and textual attention mechanisms to capture fine-grained interplay between vision and language, and that DANs attend to specific regions in images and words in text through multiple steps and gather essential information from both modalities.

\citet{fukui2016multimodal} perform fusion between visual and text modalities by computing the tensor product of the two feature vectors. Since this operation would create a very large $n^2$-sized vector for $n$-sized input vectors, several tricks are performed to compute a smaller approximation of this product, as suggested by \citet{gao2016compact}. The general architecture of the model is then set up as in most attention models, with this compact bilinear pooling used to generate both the attention maps and the final features.

\paragraph{Addition of a Knowledge Base}
\citet{wu2016ask} have proposed using an external knowledge base in a VQA system. Their approach is to obtain attributes from an image using an image labeling model (\cite{aastrom2016image}), query an external knowledge base, and use that information, along with image features generated from a CNN, to seed the initial hidden state of an LSTM. The question is then fed to the LSTM and an answer is generated.

\subsection{Image Captioning}
Image captioning is closely related to VQA: both tasks try to generate textual data from visual inputs, and captions have been used to provide supplemental information in VQA models (as in \cite{wu2016ask}). In particular, \citet{karpathy2015deep} have proposed a widely-used captioning model that is composed of two main parts. In this model, a CNN is used to obtain image region embeddings, an RNN to obtain a caption representation, and an alignment is performed between them using a Markov random field. These aligned pairs are then fed to the second model, which uses an RNN to generate a caption for each region of the image, using features computed from that region with a CNN as the initial hidden state of the RNN. \citet{xu2015show} propose another model where image features are generated with a CNN, and passed to an LSTM with attention to generate the caption.

\subsection{Counting with Neural Networks}
VQA models generally perform significantly worse on counting questions than on other types of questions (this is the case in all reviewed prior work: \cite{antol2015vqa}, \cite{shih2016where}, \cite{gao2016compact}, \cite{andreas2016neural} for instance). Yet approaches to counting with neural networks have been proposed. \citet{segui2015learning} use a straightforward model to count even digits or pedestrians in images. Their system uses two or more convolutional layers, followed by a one or several fully-connected layers.

\section{Proposed Approach}
% This is where the technical novelty of your paper should be clearly explained and motivated. 

In this work, we extend the baseline of Neural Module Networks (NMN) proposed by \cite{andreas2016neural}. 
This model is trained with triples of (question, image, answer) from which it dynamically learns to assemble a neural network in which individual modules perform specific tasks. The modules to use are selected based on the dependency parse tree\footnote{This work uses Stanford dependency parser \citep{chen2014fast} for this purpose.} of the question, and the type of the question word (i.e. the first or first few words in the question). Essentially, this combines the good performance of neural networks in image recognition and captioning with the power of classical NLP methods, by assembling the network using linguistic information.

Specifically, a question $\mathbf{Q}$ is mapped to a logical representation of meaning, from which all the nouns, verbs and prepositional phrases that are directly related to the \textit{`wh'} word (``what'', ``where'', ``how'', ``how many'', etc.) are collected. The common nouns and verbs are mapped to a \textit{find} module. The modules can then be combined using \textit{and} modules, and a \textit{measure} or a \textit{describe} module is inserted at the top. Table \ref{tab:moduleDescription} lists the roles and implementations of these modules.

\begin{table}[h]
\begin{adjustbox}{center}
\small
\begin{tabular}{ l m{8cm} ll }
 \hline
 Module & Description & Input & Output \\
\hline
Find & Convolves every position in the input image with a weight vector to produce an attention map. & Image & Attention\\
And & Merges two attention maps into a single attention map using element wise product. & Attention x Attention & Attention\\
Describe & Computes an average over image features weighted by the attention, and then passes through a single fully-connected layer. & Image x Attention & Label\\
Measure & Passes the attention through a fully connected layer, ReLU nonlinearity, fully connected layer and a softmax to generate a distribution over labels. & Attention & Label\\
\hline
\end{tabular}
\end{adjustbox}

\caption{Different NMN modules}
\label{tab:moduleDescription}
\end{table}

In any case, answer prediction is formulated as a classification problem where we are selecting the answer from 2000 most common answers that were encountered in the training dataset. This approach (with varying sizes for the number of answers considered) seems very common in prior work, and works well because the majority of answers are short and occur several times in the dataset.

The main contribution of our work is the enrichment of the features through text. We incorporate this information in three ways, two of which primarily depend on the information from captions and the third is based on the information that can be incorporated from external knowledge bases. 
The caption information that we incorporate is obtained from a pre-trained image captioning model from \citet{karpathy2015deep}, which is described in the prior art section and is trained on the MSCOCO \citep{lin2014microsoft} dataset.

We now describe the three proposed approaches in detail. In the following, let $\mathbf{q_i}$ represent the word embedding for word $i$ and $T$ be the maximum number of words in the question in that batch. Hence the entire question is represented as 
\begin{align*}
\mathbf{Q} = (\mathbf{q}_1, \mathbf{q}_2, \dots, \mathbf{q}_T)
\end{align*}
In a similar way, let the maximum number of words in a caption be $N$ and let $C_i$ represent the word embedding of word $i$ in the caption. Hence the caption is represented as 
\begin{align*}
\mathbf{C} = (\mathbf{c}_1, \mathbf{c}_2, \dots, \mathbf{c}_N).
\end{align*}

\subsection{Caption Information}
\begin{figure}[h]
	\centering
	\begin{subfigure}{0.45\linewidth}
    \centering
     \includegraphics[width=\linewidth]{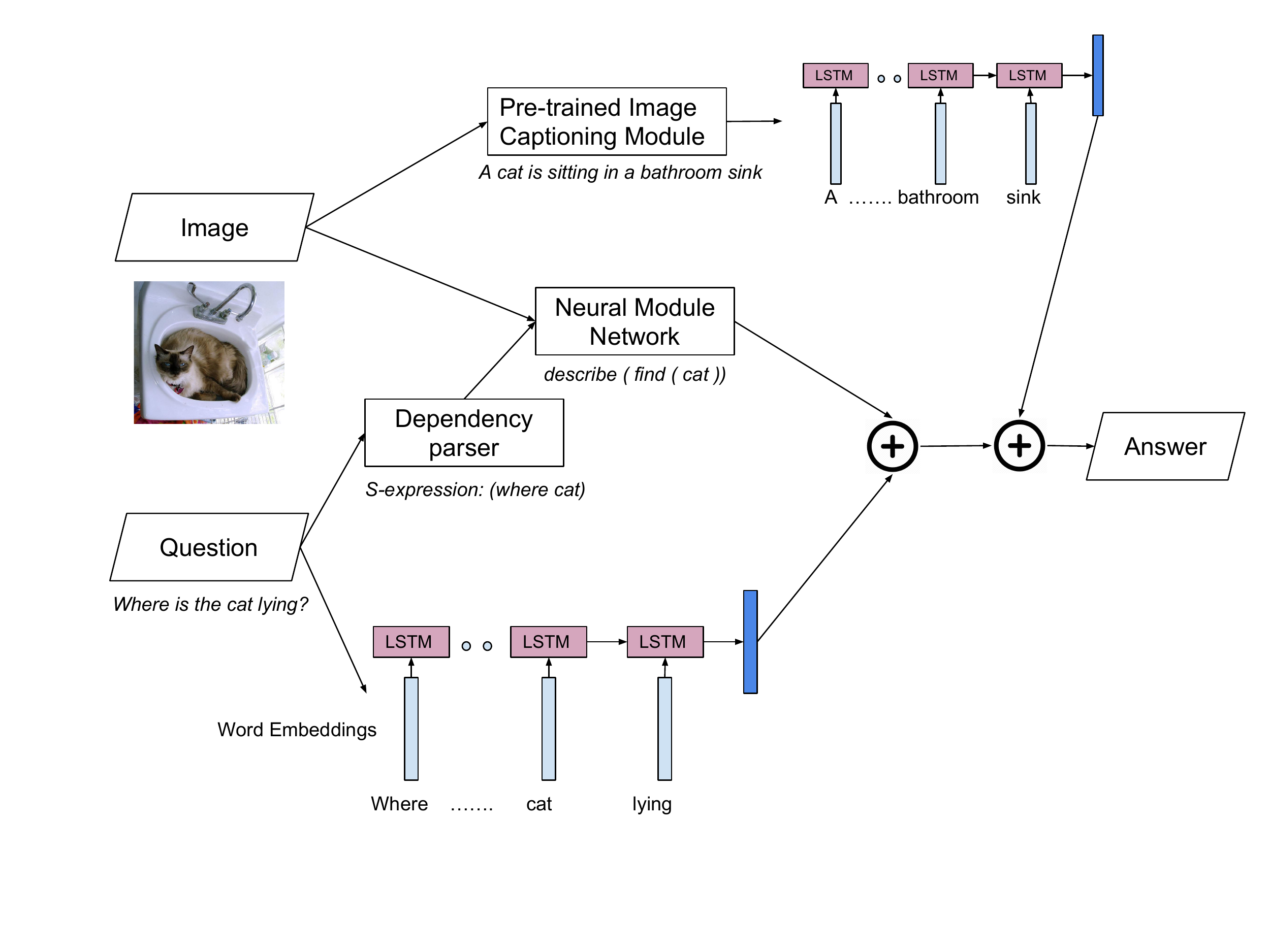}
        \caption{Textual enrichment of NMN through caption information}
        \label{fig:CaptionInformation}
    \end{subfigure}
    \hspace{1cm}
    \begin{subfigure}{0.45\linewidth}
    \centering
        \includegraphics[width=\linewidth]{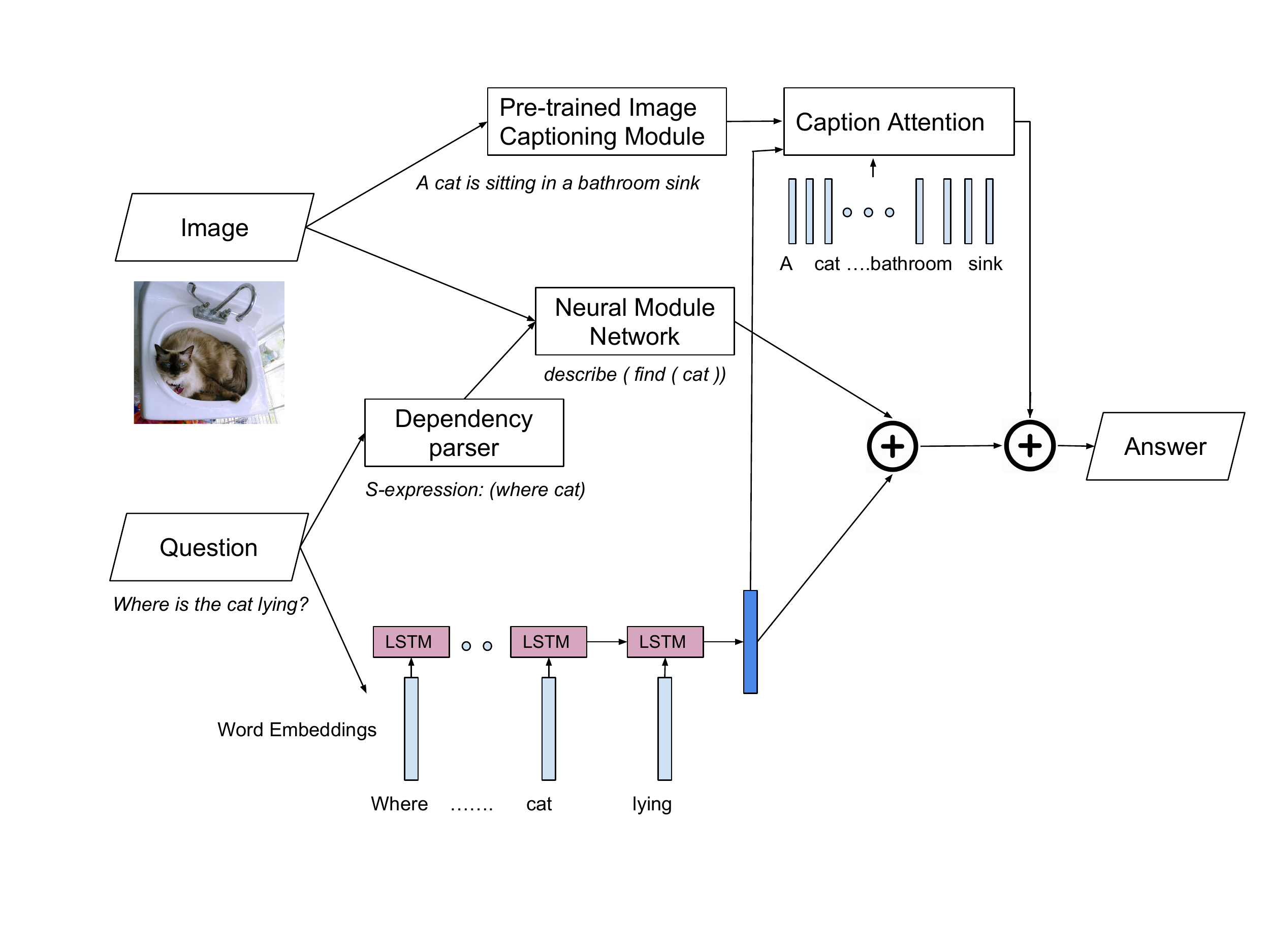}
        \caption{Textual enrichment of NMN through attention on captions}
        \label{fig:CaptionAttention}
	\end{subfigure}
    \caption{Schematic representation of our two captioning-based approaches}
\end{figure}

In this subsection, we describe the methodology of incorporating the entire caption information to assist the prediction of NMNs. Figure \ref{fig:CaptionInformation} represents the architecture of this approach.

$\mathbf{Q}$ is processed through a single layer LSTM and a fully connected layer from which a question context vector $\mathbf{m_Q}$ is obtained. The same procedure is applied to $\mathbf{C}$ using another LSTM and a fully connected layer to obtain the caption context vector $\mathbf{m_C}$. The image features along with the dependency parse are provided as input to the NMN. Let the output vector of NMN be represented as $\mathbf{Pred_{NMN}}$. An elementwise addition is performed on $\mathbf{Pred_{NMN}}$, the question context vector, and the caption context vector, followed by a rectified linear unit (ReLU) nonlinearity and finally another fully connected layer (with $\mathbf{W}$ as the weight matrix). To obtain the final answer distribution, we compute a $\mathit{softmax}$ over this vector call the output $\mathbf{Pred_{V}}$. The final answer is the word corresponding to the maximum value in this vector.
These steps are mathematically represented as the following equations:
\begin{align*}
\mathbf{Pred_{V}} &= \mathit{softmax}( \mathbf{W} * \max(0,(\mathbf{Pred_{NMN}} \oplus \mathbf{m_Q} ) \oplus \mathbf{m_C} )) \\
\mathit{answer} &= \argmax_i \hspace{0.2cm} \mathbf{Pred_{V}}[i]
\end{align*}

\subsection{Caption Attention}
In the previous approach, we notice that there are cases where the entire caption may not be helpful in answering the question being asked. Instead, an end-to-end back propagation of the error by attending to the necessary information from the caption after combining with the respective question and prediction from the NMNs can help localize on the answer space. Figure \ref{fig:CaptionAttention} outlines the architecture of this approach.

In order to capture attention on the caption words, the caption word embeddings and the question context vector are individually passed through a fully connected layer with a $\mathit{sigmoid}$ ($\sigma$) activation function. The weight matrix used for all caption word embeddings ($\mathbf{c}_i$) and question context vector ($\mathbf{m_Q}$) are $\mathbf{W_C}$ and $\mathbf{W_Q}$ respectively.
%where $\mathbf{W_C}$, $\mathbf{W_Q}$ $\in$ $\mathbb{R}^{k \times d}$ and $\mathbf{m_Q}$, $\mathbf{c}_i$ $\in$ $\mathbb{R}^{d}$
An elementwise dot product is then performed on the two resulting vectors. The hidden representation $\mathbf{H}$ is:
\begin{align*}
\mathbf{H} &= ( \mathbf{h}_1, \mathbf{h}_2, \dots, \mathbf{h}_N )  \in \mathbb{R}^{k \times N}, \text{  where } \mathbf{h}_i = \sigma (\mathbf{W_Q m_Q}) \odot \sigma (\mathbf{W_C} \mathbf{c}_i)
\end{align*}
The elementwise multiplication propagates the error that does not maximize the importance of the words in the captions with respect to the question. The projected space is thus brought back into the dimensions of the caption length and a normalized probability distribution is computed over this vector using a $softmax$ function (with $\mathbf{W_h}$ as weight parameter) to get the attention vector $\mathbf{a}$ for the captions. The caption attention context vector $\hat{\mathbf{C}}$ is then the average of the caption word embeddings weighted by the attention as shown below: 
\begin{align*}
\mathbf{ \hat{C} } &= \sum_{i=1}^{N} a_i \mathbf{c_{i}}, \hspace{0.2cm} a_i \in \mathbf{a}, \text{   where   }  
\mathbf{a} = \mathit{softmax}( \mathbf{W_h H} )
\end{align*}
$\mathbf{Pred_{NMN}}$, $\mathbf{m_Q}$, and the caption attention context vector $\mathbf{\hat{C}}$ are added element wise, followed by ReLU nonlinearity and a fully connected layer (with $\mathbf{W}$ as weight parameter) over which we obtain the probability distribution in the answer space using a $\mathit{softmax}$ function, as before.
\begin{align*}
\mathbf{Pred_{V}} &= \mathit{softmax}( \mathbf{W} * \max(0, (\mathbf{Pred_{NMN}} \oplus \mathbf{m_{Q}} ) \oplus \mathbf{\hat{C}} )) \\
\mathit{answer} &= \argmax_i \hspace{0.2cm} \mathbf{Pred_{V}}[i]\\
\end{align*}

\subsection{External Knowledge Sources}

% remind people how your proposed model/approach addresses the challenges highlighted in the
% introduction. Also, make sure that the novelty of your approach is highlighted (i.e., what part was already done vs. what part is novel).

% Model Description: formalize mathematically your proposed model or approach. This
% should include the mathematical definition of the variables involved in your problem
% (e.g., input variable x and labels y). You should also include an objective function
% describing what you are modeling (e.g., P(y|x)). While this section may be relatively
% short (e.g., half a page), you should be consistent with your notation later in the paper.

\begin{figure}[h]
\begin{adjustbox}{center}
    \begin{subfigure}[b]{0.5\textwidth}
    	\tiny\begin{tabular}{lr}
        \begin{minipage}{0.5\textwidth}
        \includegraphics[width=\textwidth]{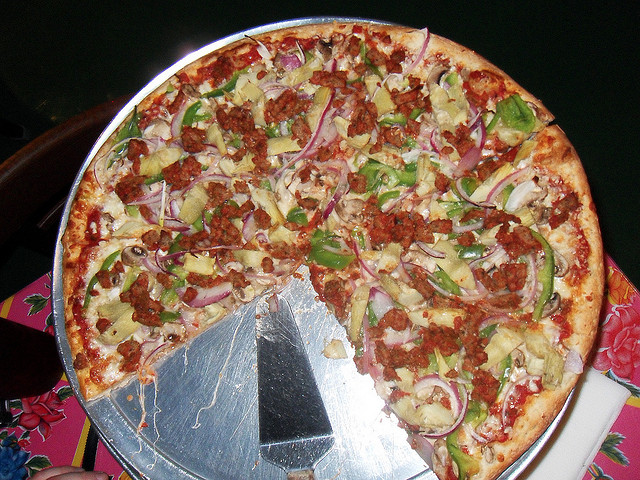}
        \end{minipage} &
        \begin{minipage}{0.5\textwidth}
        \textbf{Question}: What toppings are on the pizza? \\
        \textbf{Answers}: meat; onion; peppers \\
        \textbf{Excerpt from Wikipedia article ``pizza''}: It is commonly topped with a selection of meats, vegetables and condiments.
        \end{minipage}
        \end{tabular}
        \caption{}
        \label{fig:pizza}
    \end{subfigure}
    ~ %add desired spacing between images, e. g. ~, \quad, \qquad, \hfill etc. 
    \hspace{1cm}
    \begin{subfigure}[b]{0.5\textwidth}
    	\tiny\begin{tabular}{lr}
        \begin{minipage}{0.5\textwidth}
        \includegraphics[width=\textwidth]{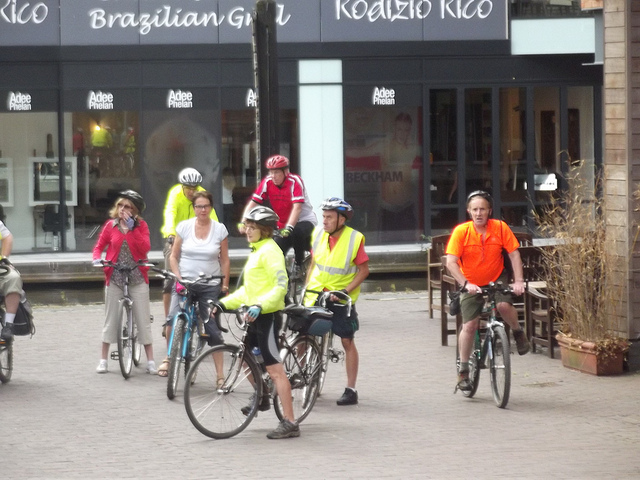}
        \end{minipage} &
        \begin{minipage}{0.5\textwidth}
        \textbf{Question}: Why are the people wearing helmets? \\
        \textbf{Answers (top 3)}: so they don't injure their heads; for protection; safety \\
        \textbf{Excerpt from Wikipedia article ``helmet''}: A helmet is a form of protective gear worn to protect the head from injuries.
        \end{minipage}
        \end{tabular}
        \caption{}
        \label{fig:helmet}
    \end{subfigure}
    \end{adjustbox}
    \caption{Examples of images that could be helped by the addition of an external knowledge source}
    \label{fig:ExternalQuestions}
\end{figure}

% Learning: Describe how you are learning the parameters of your model. State clearly
% your loss function and the learning algorithm used in your experiments. If you have
% enough space, give derivation of your gradient, at least for some of the main model
% parameters.
A fraction of the questions in the dataset seem to call for some external general knowledge. Two examples are given in figure \ref{fig:ExternalQuestions}. In \ref{fig:pizza}, the top answer (``meat'') is present in the Wikipedia abstract of the article titled ``Pizza''. In \ref{fig:helmet}, one can easily argue that the image is not necessary to answer the question, while access to information about what a helmet is useful for giving a direct answer. While it is difficult to objectively evaluate how many questions can benefit from the addition of an external knowledge source, a cursory look at 100 images from the VQA 1.0 dataset suggests that about one image in every fifteen would be helped by the addition of general knowledge.

To alleviate this problem, we add support for external knowledge bases in NMN. As suggested by \citet{wu2016ask}, we use information extracted from our knowledge base as the seed for the hidden state of the LSTM that will parse the question. While it is known that LSTMs tend to forget their initial state if their input is too long (\cite{neubig2016} for instance), we do not believe this to be an issue here since questions are generally under ten words in length (3\% of all questions exceed this length). 

The novelty of our approach essentially resides in the selection and the preprocessing of the knowledge base. We use the DBPedia collection of English Wikipedia abstracts \citep{auer2007dbpedia} as our knowledge base. Articles are filtered to keep only those that correspond to proper or common nouns, in order to remove the many observed false matches on movie or song titles. This allows us to match articles in the content of the abstract, instead of the title or meta-information only, potentially giving access to more information if no direct matches are found, as well as being able to take in more information, since Wikipedia abstracts are more detailed than the DBPedia ontology elements used by \citet{wu2016ask}. A potential drawback of this approach is that longer abstracts may not be harder to exploit as expected if they contain more information that is irrelevant to the question. Essentially, our hypothesis for this part is that giving the model access to more external information should be beneficial. The entire collection of abstracts is indexed in an Apache Lucene\footnote{\url{https://lucene.apache.org/}} database. Each question is then parsed to select only nouns using NLTK (Natural Language Toolkit) \citep{loper2002nltk}, and a query is run against the database using the nouns extracted from the question. All selected articles are then turned into 300-dimensional vectors using Doc2Vec \citep{le2014distributed}. We use a pre-trained model from \citet{lau2016empirical} for this task.

\section{Experimental Setup}

We evaluate the performance of our model on the VQA 1.0 dataset \citep{antol2015vqa}. This dataset is widely used for VQA tasks, comprising of around 200,000 images from
MSCOCO \cite{lin2014microsoft}. Each of these images is associated with three different questions along with ten answers to each of these questions which were generated by human annotators. We train our model using the standard train/val/test split. The \texttt{conv5} layer after performing max pooling, from a 16-layer VGGNet (a deep CNN, \cite{simonyan14deep}) is used generate the visual features, that are normalized to have a mean of 0 and a standard deviation of 1, used by NMN. 

We use ApolloCaffe \citep{jia2014caffe} to develop our model.  Since we treat the problem as a classification problem, categorical cross entropy is used as a loss function. We use the ADADELTA optimizer with standard parameter settings \citep{zeiler2012adadelta}. We set a batch size to be 100 and train up to 12 epochs with early stopping if the validation accuracy has not improved. In Caption Attention Model, the hidden layer size of $\mathbf{W_Q}$ and $\mathbf{W_C}$ is set to 200. The question context vector is generated through a single layer LSTM of 1000 hidden units and the caption context vector in Caption Attention Model is generated through a single layer LSTM of 500 hidden units.
%The dimensions of the question and caption context vectors built from the LSTM are 1000 and 500 respectively.

\section{Results and Discussion}

The two baselines we explore are hierarchical co-attention models \citep{lu2016hierarchical} and NMNs \citep{andreas2016neural}. Table \ref{tab:baselineComparison} shows the results that are replicated by training these models with train+val as the training set.

\begin{table}[h]
\begin{adjustbox}{center}
\begin{tabular}{ lcccc } 
 \hline
 Models & Yes/No & Number & Other & Overall \\
\hline
Hierarchical Question-Image Co-Attention & 79.6 & 38.4 & 49.1 & 60.5 \\
Neural Module Networks & 80.8 & 36.1 & 43.7 & 58.1 \\
Neural Module Networks (Reported in paper) & 81.2 & 35.2 & 43.3 & 58.0 \\% & & & \\
\hline
\end{tabular}
\end{adjustbox}
\caption{Comparison of baseline models on test-dev open-ended questions. The published implementation of NMN2 performs somewhat better than reported in the paper, and both do worse than Hierarchical Co-Attention.}
\label{tab:baselineComparison}
\end{table}

The major focus of our work lies in improving the NMNs by combining some advantages from the attention model. Henceforth, we present the results of our system obtained by improving the baseline of NMNs. The reason for this choice though it is not the state of the art model, is the intuitive and interesting combination of dependency trees to dynamically adapt the neural networks by assembling different neural modules.

\subsection{Experimental results}
\begin{table}[h]
\begin{adjustbox}{center}
\begin{tabular}{ lcccccc } 
 \hline
Model Name & Train Acc & Validation Acc & Yes/No & Number & Other & Overall \\
\hline
Neural Module Networks (NMN) & 60.0 & 55.2 & 79.0 & 37.5 & 42.4 & 56.9 \\
Caption Attention Only & 50.0 & 48.2 & 78.4 & 36.5 & 27.8 & 49.5 \\
\textbf{NMN + Caption Information} & \textbf{61.9} & \textbf{56.4} & \textbf{79.8} & \textbf{37.4} & \textbf{42.1} & \textbf{57.1} \\
NMN + Caption Attention & 60.3 & 55.2 & 79.2 & 35.8 & 42.1 & 56.6 \\
NMN + External Knowledge Source & 61.3 & - & 79.2 & 36.4 & 42.2 & 56.8 \\ 
\hline
\end{tabular}
\end{adjustbox}
\caption{Comparison of our different models and the baseline on test-dev open-ended questions}
\label{tab:ourApprochesResults}
\end{table}

Table \ref{tab:ourApprochesResults} uses train set whereas table \ref{tab:measureParse} uses train+val as the training set.
As we observe in table \ref{tab:ourApprochesResults}, the train and test accuracy in the second row, which corresponds to the experiment by removing the output from the NMN completely and just relying on the attentions from the captions is about 50\%. This indicates that the model is able to learn to predict some answers from the captions to the images. 

Adding caption information increases the train and test accuracy by 1.6 percentage points and 1.2 points respectively. However, our caption attention model degrades NMN performance. One possible reason for this could be that we are forcing the model to attend to some words in the caption in cases where the caption is not relevant to the question being asked.

%Some general observations that we made during this process are that the dataset has high number of `yes/no' questions, majority of the answers for such questions being `yes'. Training beyond 6 iterations is overfitting the model. This leads to the increase in the overall accuracy which is biased towards predicting `yes' but does not perform well in other question categories.

To combat information loss when the question parse is simplified into a network layout, we have experimented with using larger parse trees. For example consider the question, `What is this person playing?'. The shorter version of the parse that baseline NMN uses is  `\textit{Describe(Find(Person))}. This parse does not include any information about `playing'. Hence we consider tree from longer parse \textit{Describe(And(Find(Person), Find(Playing)))}. In addition, we implement a \textbf{\textit{measure}} module to address counting type questions specifically\footnote{The \textit{measure} module is described in \cite{andreas2016neural} but is not implemented in the published code. We implement this module as outlined in the paper; it is very similar to the counting system in \citet{segui2015learning}}.

\begin{table}[h]
\begin{center}
\begin{tabular}{ lccccc } 
 \hline
Model Name & Train Acc & Yes/No & Number & Other & Overall \\
\hline
Neural Module Networks (NMN) & 62.9 & 80.8 & 36.1 & 43.7& 58.1 \\
Longest Parse & 61.4 & 79.7 & 35.8 & 42.3 & 57.0 \\
Measure Module & 64.0 & 79.6 & 34.8 & 41.7 & 56.5 \\
\hline
\end{tabular}
\end{center}
\caption{Comparison of results using some simple tweaks to the NMN model. All of our changes lead to worse results, so we do not use them for our other experiments.}
\label{tab:measureParse}
\end{table}

As we can see from table \ref{tab:measureParse}, performing less simplification on the parse tree decreases the scores. This could be due to a higher chance of error in complex parses, which are consequently not mapped to appropriate neural modules. 

Adding a \textit{measure} module at the top of the parse tree instead of a \textit{describe} module for counting questions (those that start with ``how many'') does not improve the results either. This could be due to the errors in the attention maps from the \textit{find} module, or to the fact that while the \textit{describe} module takes both image and attention maps as inputs, the \textit{measure} module only takes the attention maps, so some information from image data might have been lost.

\subsection{Comparative Qualitative Analysis}
To analyze the performance of our approach  with respect to the baseline of NMNs, we randomly select some images from the validation set and compare the answers produced by our system with correct answers and those produced with NMNs. We present this analysis in table \ref{tab:randomExamples}. The notations `Q' and `C' represent corresponding questions and captions. ``Predicted Answer NMN'' is the output from the NMN model and ``Predicted Answer NMN+CA'' is the answer predicted by our model after adding the attention on captions with respect to the question. 

\begin{table}[h]
\centering
{\tiny
\begin{tabular}{p{4cm}p{4cm}p{4cm}}
\includegraphics[width=0.3\textwidth]
{COCO_val2014_000000246425} & \includegraphics[width=0.20\textwidth]{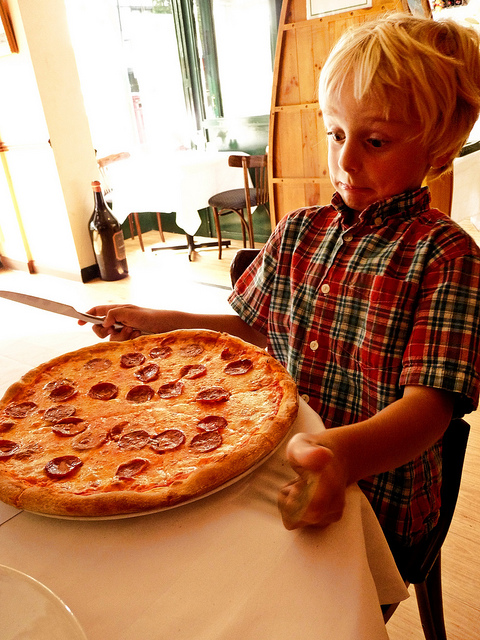} & \includegraphics[width=0.2\textwidth]{COCO_val2014_000000080172}\\
 Q: How many planes are flying? &
 Q: What is the food called? &
 Q: What is the child holding?\\
 C: two planes flying in the sky in the sky &
 C: a little boy sitting at a table with a pizza &
 C: a young boy brushing his teeth in the bathroom \\
 Correct Answer: 2 &
 Correct Answer: pizza&
 Correct Answer: toothbrush\\
 Predicted Answer NMN: 4&
 Predicted Answer NMN: pizza&
 Predicted Answer NMN: phone\\
 Predicted Answer NMN+CA: 2&
 Predicted Answer NMN+CA: pizza&
 Predicted Answer NMN+CA: toothbrush\\
 (a) & (b) & (c) \\ 
\includegraphics[width=0.26\textwidth]
{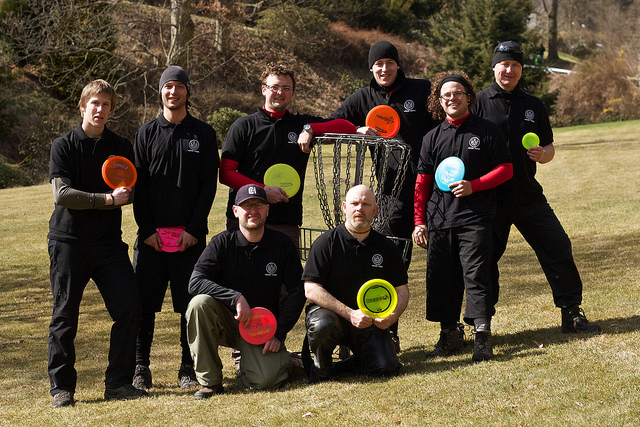} & \includegraphics[width=0.26\textwidth]{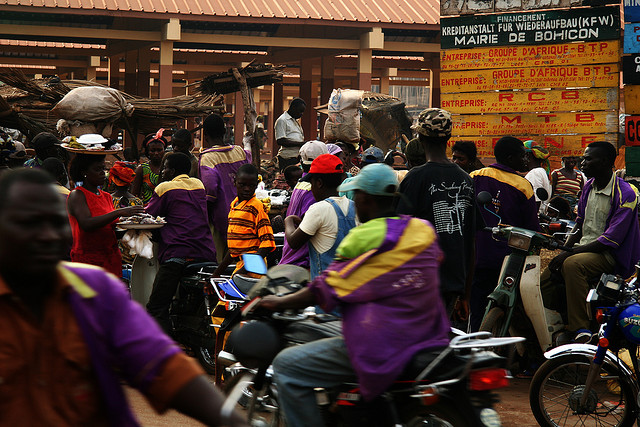} & \includegraphics[width=0.26\textwidth]{COCO_val2014_000000413247}\\
Q: How many people are standing? &
Q: What color is the sign?&
Q: Is there a tree on the desk?\\
C: a group of people in a field with a frisbee &
C: a group of people riding motorcycles down a street &
C: a laptop computer sitting on top of a desk\\
Correct Answer: 6&
Correct Answer: black and white&
Correct Answer: no\\
Predicted Answer NMN: 2&
Predicted Answer NMN: red&
Predicted Answer NMN: yes\\
Predicted Answer NMN+CA: 5&
Predicted Answer NMN+CA: red&
Predicted Answer NMN+CA: yes\\
(d) & (e) & (f) 
\end{tabular}}
\caption{Comparative performance analysis of before and after adding Caption Attention to the NMNs. The first row of examples show the questions correctly by our model. The second row of examples show questions where both NMN and our model predict the wrong answer.}
\label{tab:randomExamples}
\end{table}
In example (a), the information about `two planes' is clearly mentioned in the caption and hence this additional information helps our model capture this answer from the embedded context vector when the caption is passed through an LSTM. Similarly the answer words `pizza' and `tooth brush' in (b) and (c) are explicitly present in the caption. An obvious drawback to our approach is when the caption produced for the image is not relevant to the question being asked.

For the examples in the second row, the captions are not exactly relevant to the question and hence attention over the captions has not been particularly helpful. The predicted answers by our model exactly match that from NMN in (e) and (f). We observed an interesting type of error that our model makes by mapping generic terms to frequent occurrences in the captions. This is depicted in (d). The caption for this image has the word `a group of people' that correspond to the number of people in the image. While NMN predicted the answer as 2, the model with caption attentions predicted the answer as 5. This could be due to frequent associations mapping the word in the caption `group' to the answer `5' in the training data and this is learnt by our model during end-to-end error backpropagation. Though the final predicted answer is wrong, common sense information that `two people' are not called group and a relatively higher number is required to be called a group is learnt by the model. 

\section{Conclusion and Future Directions}
% ending of Section2

We show that incorporating the information from captions improves results slightly (by 1.9\% on training and by 1.2\% on testing), especially in cases where the caption is relevant to the question being asked. However, our model fails when the generated caption is not relevant to the question and hence one future direction is towards generating captions with certain required words. This approach also learns appropriate mappings to generic terms mapped to more probable associations from the cations in the training data. An interesting direction to explore in the future would be the analysis of the irrelevant captions generated for an image for better fitting attention models. 

%\clearpage
% ending here
\subsubsection*{Acknowledgments}
We thank Professor Louis-Phillipe Morency, Dr. Tadas Baltrusaitis, Amir Zadeh and Chaitanya Ahuja for providing us the guidance, constant timely feedback and resources required for this project.

% \nocite{*}
\small
\bibliographystyle{apalike}
\bibliography{ref}

\end{document}